\pdfoutput=1

\documentclass[11pt]{article}

\usepackage{emnlp2021}

\usepackage{times}
\usepackage{cleveref}
\usepackage{latexsym}
\usepackage{graphicx}
\usepackage{linguex}
\usepackage{subcaption}

\usepackage[T1]{fontenc}

\usepackage[utf8]{inputenc}

\usepackage{microtype}

\usepackage{footnote}
\usepackage{footmisc}

\renewcommand{\thefootnote}{\alph{footnote}}

\newcommand{\astfootnote}[1]{%
\let\oldthefootnote=\thefootnote%
\setcounter{footnote}{0}%
\renewcommand{\thefootnote}{\fnsymbol{footnote}}%
\footnote{#1}%
\let\thefootnote=\oldthefootnote%
}

\newcommand{\sysname}{DisCoDisCo}

\title{\sysname{} at the DISRPT2021 Shared Task: A System for \\ Discourse Segmentation, Classification, and Connective Detection}

\author{
    Luke Gessler,
    Shabnam Behzad,
    Yang Janet Liu,
    Siyao Peng,
    Yilun Zhu,
    Amir Zeldes \\
    Corpling Lab \\
    Georgetown University \\
    \{lg876, 
    sb1796, 
    yl879,
    sp1184, 
    yz565, 
    az364\}@georgetown.edu\\
}

\begin{document}
\maketitle
\begin{abstract}
This paper describes our submission to the DISRPT2021 Shared Task on Discourse Unit Segmentation, Connective Detection, and Relation Classification. 
Our system, called \sysname{}, is a Transformer-based neural classifier which enhances contextualized word embeddings (CWEs) with hand-crafted features, relying on tokenwise sequence tagging for discourse segmentation and connective detection, and a feature-rich, encoder-less sentence pair classifier for relation classification. 
Our results for the first two tasks outperform SOTA scores from the previous 2019 shared task, and results on relation classification suggest strong performance on the new 2021 benchmark. 
Ablation tests show that including features beyond CWEs are helpful for both tasks, and a partial evaluation of multiple pre-trained Transformer-based language models indicates that models pre-trained on the Next Sentence Prediction (NSP) task are optimal for relation classification.\astfootnote{Full disclosure: our team includes both organizers and dataset annotators from the shared task. All code is available at \url{https://github.com/gucorpling/DisCoDisCo}}
\end{abstract}

\section{Introduction}

Recent years have seen tremendous advances in NLP systems' ability to handle discourse level phenomena, including discourse unit segmentation and connective detection \cite{zeldes-etal-2019-disrpt} as well as discourse relation classification (e.g.~\citealt{lin_ng_kan_2014, braud-etal-2017-cross, kobayashi-etal-2021-improving}). 
For segmentation and connective detection, the current state of the art (SOTA) is provided by models using Transformer-based, pretrained contextualized word embeddings \cite{MullerBraudMorey2019}, focusing on large context windows without implementation of hand-crafted features. 
For relation classification, SOTA performance on the English RST-DT benchmark \cite{CarlsonEtAl2003} has been achieved by neural approaches \citep{guz-carenini-2020-coreference, Kobayashi_Hirao_Kamigaito_Okumura_Nagata_2020, nguyen-etal-2021-rst, kobayashi-etal-2021-improving}. 
For PDTB-style data, the 2015 and 2016 CoNLL shared tasks on shallow discourse parsing \cite{xue-etal-2015-conll, xue-etal-2016-conll} have motivated work on both explicit (e.g.~\citealt{kido-aizawa-2016-discourse}) and implicit (e.g.~\citealt{liu-etal-aaai-implicit, wang-lan-2016-two, rutherford-etal-2017-systematic, kim-etal-2020-implicit, liang-etal-2020-extending, zhang-etal-2021-context}) discourse relation classification in English PDTB-2 \cite{PrasadEtAl2008} and PDTB-3 \cite{PrasadWebberLeeEtAl2019} as well as on the PDTB-style Chinese newswire corpus (CDTB, \citealt{zhou-xue-2012-pdtb, cdtb-zhou-etal}) such as \citet{schenk-etal-2016-really} and \citet{weiss-bajec-2016-discourse}.

Our system for DISRPT 2021, called \sysname{},
({\bf Dis}trict of {\bf Co}lumbia {\bf Dis}course {\bf Co}gnoscente)
extends the current SOTA architecture by introducing hand-crafted categorical and numerical features that represent salient aspects of documents' structural and linguistic properties. 
While Transformer-based contextualized word embeddings (CWEs) have proven to be rich in linguistic features, they are not perfect \citep{rogers-etal-2020-primer}, and there are some textual features---such as position of a sentence within a document, or the number of identical words occurring in two discourse units---which are difficult or impossible for a typical Transformer-based CWE model to know.
We therefore supplement CWEs with hand-crafted features in our model, with special attention paid to features we expect CWEs to have a poor grasp of.

We implement our system with a pretrained Transformer-based contextualized word embedding model at its core, and dense embeddings of our hand-crafted features incorporated into it.
Our exact approach varies by task: we use a tokenwise classification approach for EDU segmentation, a CRF-based sequence tagger for connective detection, and a BERT pooling classifier for relation classification.  
Our system is implemented in PyTorch \citep{pytorch} using the framework AllenNLP \cite{GardnerEtAl2018}. 
Our results show SOTA scores exceeding comparable numbers from the 2019 shared task, and ablation studies indicate the contribution of features beyond CWEs.

\section{Previous Work}

\paragraph{Segmentation and Connective Detection} Following the era of rule-based segmenters (e.g. \citealt{Marcu2000}, \citealt{ThanhAbeysingheHuyck2004}), \citet{SoricutMarcu2003} used probabilistic models over constituent trees for token-wise binary classification (i.e.~boundary/no-boundary). \citet{SporlederLapata2005} used a two-level stacked boosting classifier on chunks, POS tags, tokens and sentence lengths, among other features. \citet{HernaultPrendingerEtAl2010} used an SVM over token and POS trigrams as well as phrase structure trees. 

More recently, \citet{BraudLacroixSoegaard2017} used bi-LSTM-CRF sequence labeling on dependency parses, with words, POS tags, dependency relations, parent, grandparent, and dependency direction, achieving an F$_1$ of 89.5 on the English RST-DT benchmark \cite{CarlsonEtAl2003} with parser-predicted syntax. Approaches using CWEs as the only input feature \cite{MullerBraudMorey2019} have achieved an F$_1$ of 96.04 on the same dataset with gold sentence splits and 93.43 without, while for some smaller English and non-English datasets, approaches incorporating features and word embeddings remain superior (e.g.~for English STAC and GUM, as well as Dutch RST data, \citealt{YuZhuLiuEtAl2019}; and for Chinese, \citealt{bourgonje-schafer-2019-multi}; for more on these datasets see below).

For connective detection, \citet{pitler2009using} used a MaxEnt classifier with syntactic features extracted from gold Penn Treebank \cite{MarcusSantoriniMarcinkiewicz1993} parses of PDTB \cite{PrasadEtAl2008} articles. \citet{patterson2013predicting} presented a logistic regression model trained on eight relation types from PDTB, with features in three categories: \textit{Relation-level} such as the connective signaling the relation; \textit{Argument-level} such as size or complexity of argument spans; and \textit{Discourse-level} features, targeting dependencies between the relation and its neighboring relations in the text (cf.~our approach to featurizing overall utilization of argument spans in the data below). \citet{polepalli2012automatic} used SVM and CRF to identify connectives in biomedical texts  \cite{prasad2011biomedical}, with features such as POS tags, dependencies and domain-specific semantic features included several biomedical gene/species taggers, in addition to predicted biomedical NER features. 

Current SOTA approaches rely on sequence labeling in a BIO scheme with CWEs from either plain text \cite{MullerBraudMorey2019} or integrating word embeddings and dependency tree features (POS, dependencies, phrase spans, \citealt{YuZhuLiuEtAl2019}), dependending on the dataset and availability of gold standard features.

\paragraph{Discourse Relation Classification} Generally speaking, discourse relation classification assigns a relation label to two pieces of texts from a set of predefined coherence or rhetorical relation labels \cite{stede2011discourse}, which varies across different discourse frameworks, corpora, and languages. Given different perspectives and theoretical frameworks, the implementation and evaluation of the relation classification task varies considerably. 

In Rhetorical Structure Theory (RST, \citealt{mann1988rhetorical}), discourse relations hold between spans of text and are hierarchically represented in a tree structure \cite{Zeldes2018book}. Performance is evaluated and reported using the micro-averaged, standard Parseval scores for a binary tree representation, following \citet{morey-etal-2017-much}. Current SOTA performance \cite{kobayashi-etal-2021-improving} on the English RST-DT benchmark \cite{CarlsonEtAl2003} with gold segmentation achieved a micro-averaged original Parseval score of 54.1 by utilizing both a span-based neural parser \cite{Kobayashi_Hirao_Kamigaito_Okumura_Nagata_2020} and a two-staged transition-based SVM parser \cite{wang-etal-2017-two} as well as leveraging silver data. 

Since PDTB is a lexically grounded framework, discourse relation classification is also called sense classification in PDTB-style discourse parsing: a sense label is assigned to the discourse connective between two text spans when a discourse connective is present (i.e.~explicit relation classification) or a label is assigned to an adjacent pair of sentences when no discourse connective is present (i.e.~implicit relation classification) \cite{slp3chapter22}. Explicit relation classification is easier as the presence of the connective itself is considered the best signal of the relation label. Most systems from the 2016 CoNLL shared task on shallow discourse parsing adopted machine learning techniques such as SVM and MaxEnt with hand-crafted features \cite{xue-etal-2016-conll}. For instance, for the English PDTB-2 \cite{PrasadEtAl2008}, \citet{kido-aizawa-2016-discourse} achieved the best performance (an F$_1$ = 90.22) on the explicit relation classification task by implementing a majority classifier and a MaxEnt classifier while \citet{wang-lan-2016-two} achieved the best performance (F$_1$ = 40.91) on implicit relation classification using a convolutional neural network. \citet{wang-lan-2016-two} also achieved the best performance on the Chinese CDTB dataset \cite{zhou-xue-2012-pdtb} in the implicit relation classification task. 

More recent work on implicit relation classification has adopted a graph-based context tracking network to model the necessary context for interpreting the discourse and has gained better performance on PDTB-2 \cite{zhang-etal-2021-context}. In addition, the increase in the number of implicit relation instances in PDTB-3 \cite{PrasadWebberLeeEtAl2019} has sparked more interest in exploring their recognition, such as \citet{kim-etal-2020-implicit} and \citet{liang-etal-2020-extending}. \citet{kim-etal-2020-implicit} presented the first set of results on implicit relation classification for both top-level senses (four labels) and more fined-grained level-2 senses (amounting to 11 labels) in PDTB-3 from two strong sentence encoder models using BERT \cite{devlin-etal-2019-bert} and XLNet \cite{NEURIPS2019_dc6a7e65-xlnet}. Due to the novelty of the DISRPT 2021 relation classification task, which combines implicit and explicit relation classification across frameworks for an unlabeled graph structure, comparable scores do not yet exist at the time of writing.

\section{Approach}

Our system comprises two main components: one targeting segmentation and connective detection using neural sequence tagging (as binary classification and BIO tagging respectively), and one targeting relation classification using BERT \citep{devlin-etal-2019-bert} fine-tuning.
We further enhance both components with the use of hand-crafted categorical and numeric features by encoding them in dense embeddings and introducing them into our neural models.

\subsection{Segmentation and Connective Detection}

Our model for segmentation and connective detection is structured as a sequence tagging model, as might be used for a task like POS tagging or entity recognition: the text is embedded, encoded with a single bi-LSTM, and decoded.

\newcommand\smallmodel[1]{{\tt\scriptsize #1}}
\newcommand\model[1]{{\tt\footnotesize #1}}
\newcommand\twoline[2]{\begin{tabular}{@{}c@{}}#1 \\ #2\end{tabular}}
\begin{table*}[h!tb]
\centering
\begin{tabular}{lll}
\hline
\textbf{Lng.} & \textbf{Segmentation/Connective Detection} & \textbf{Relation Classification} \\ 
\hline
\smallmodel{deu} & \smallmodel{xlm-roberta-large} & \smallmodel{bert-base-german-cased} \\
\smallmodel{eng} & \smallmodel{google/electra-large-discriminator} & \smallmodel{bert-base-cased} \\
\smallmodel{eus} & \smallmodel{ixa-ehu/berteus-base-cased} & \smallmodel{ixa-ehu/berteus-base-cased} \\
\smallmodel{fas} & \smallmodel{HooshvareLab/bert-fa-base-uncased} & \smallmodel{HooshvareLab/bert-fa-base-uncased} \\
\smallmodel{fra} & \smallmodel{xlm-roberta-large} & \smallmodel{dbmdz/bert-base-french-europeana-cased} \\
\smallmodel{nld} & \smallmodel{pdelobelle/robbert-v2-dutch-base} & \smallmodel{GroNLP/bert-base-dutch-cased} \\
\smallmodel{por} & \smallmodel{neuralmind/bert-base-portuguese-cased} & \smallmodel{neuralmind/bert-base-portuguese-cased} \\
\smallmodel{rus} & \smallmodel{DeepPavlov/rubert-base-cased} & \smallmodel{DeepPavlov/rubert-base-cased-sentence} \\
\smallmodel{spa} & \smallmodel{dccuchile/bert-base-spanish-wwm-cased} & \smallmodel{dccuchile/bert-base-spanish-wwm-cased} \\
\smallmodel{tur} & \smallmodel{dbmdz/bert-base-turkish-cased} & \smallmodel{dbmdz/bert-base-turkish-cased} \\
\smallmodel{zho} & \smallmodel{bert-base-chinese} & \smallmodel{hfl/chinese-bert-wwm-ext} \\
\hline
\end{tabular}
\caption{\label{tab:lms} CWE Models used, by language. All models were obtained from \url{huggingface.co}'s registry. Note that there is one exception for relation classification: on \model{eng.sdrt.stac}, \model{bert-base-uncased} is used.}
\end{table*}

\begin{table*}[ht]
    \centering
    \begin{tabular}{llll}
    \hline
        \textbf{Feature} & \textbf{Type} & \textbf{Example} & \textbf{Description}  \\\hline
        UPOS tag & Cat. & PROPN & UD POS tag \\
        XPOS tag & Cat. & NNP & Language-specific POS tag \\
        UD deprel & Cat. & advmod & UD dependency relation \\
        Head distance & Num. & 5 & Distance from a word to its head in its UD tree \\
        Sentence type & Cat. & subjunctive & Captures mood and other high-level sentential features \\ 
        Genre & Cat. & reddit & Genre of a document (where available, as in \model{eng.rst.gum}) \\ 
        Sentence length & Num. & 23 & Length, in tokens, of a sentence. \\
    \hline
    \end{tabular}
    \caption{Summary of 7 of the 12 features used for the segmentation and connective detection module. Every categorical feature is embedded in a space whose size is the square root of the total number of labels for the feature, and numerical features are log scaled.}
    \label{tab:segfeats}
\end{table*}

In the embedding layer, we rely on three kinds of embeddings: bi-LSTM encoded character embeddings ($d$ = 64); language-specific fastText \citep{bojanowski-etal-2017-enriching} static word embeddings ($d$ = 300); and language-specific contextualized word embeddings from pretrained models posted publicly on HuggingFace's model registry at \url{huggingface.co} and used via HuggingFace's \texttt{transformers} library \citep{wolf-etal-2020-transformers} ($d$ = 768/1024).
The fastText embeddings are kept frozen during training, but the pretrained Transformer model's parameters are trainable, at a lower learning rate.
Average pooling is used to obtain word-level representations from CWE sub-word representations. 
Multiple CWEs were evaluated for each language, and the one that yielded the best performance on the validation splits of the corpora for that language was selected, shown in \Cref{tab:lms}.
These three representations are concatenated, yielding a vector of size $d_{\mathrm{emb}}$ for each word.

In the next layer, we encode the embeddings along with a variety of features and a representation of the preceding and following sentence (see below).
The features we compute are tokenwise, and cover a variety of grammatical and textual information that we expected would be useful for the task. 
Some of the features are described in \Cref{tab:segfeats}. 
In order to convert these features into tensors, every categorical feature is embedded in a space as big as the square root of the total number of labels for the feature, and every numerical feature is log scaled. 
This yields an additional $d_{\mathrm{feat}}$ dimensions for each word. 

In addition to the features, we also compute a representation of the current sentence's two neighboring sentences by embedding them and using a bi-LSTM to summarize them into a relatively low-dimensional ($d_{\mathrm{neighbors}}$ = 400) vector, which is concatenated onto every word's vector.
Combining the feature dimensions and the neighboring sentences' dimensions, our input to the encoder is of size $d_{\mathrm{enc}} = d_{\mathrm{emb}} + d_{\mathrm{feat}} + d_{\mathrm{neighbors}}$.

The sequence is fed through a bi-LSTM, and the label for each token is then predicted either by a linear projection layer or conditional random fields: CRF is used for connective detection datasets, and the linear projection layer is used for segmentation datasets.\footnote{We initially used a CRF on all datasets, but our experiments showed a small degradation on segmentation datasets when using CRF.}

\begin{table*}[ht]
    \centering
    \begin{tabular}{llll}
    \hline
        \textbf{Feature} & \textbf{Type} & \textbf{Example} & \textbf{Description}  \\\hline
        Genre & Cat. & reddit & Genre of a document (where available, as in eng.rst.gum) \\ 
        Children* & Num. & 2 & No. of child discourse units each unit in the pair has \\ 
        Discontinuous* & Cat. & false & Whether the unit's tokens are not all contiguous in the text \\
        Is Sentence* & Cat. & true & Whether the unit is a whole sentence \\ 
        Length Ratio & Num. & 0.3 & Ratio of unit 1 and unit 2's token lengths \\ 
        Same Speaker & Cat. & true & Whether the same speaker produced unit 1 and unit 2 \\
        Doc. Length & Num. & 214 & Length of the document, in tokens \\ 
        Position* & Num. & 0.4 & Position of the unit in the document, between 0.0 and 1.0 \\
        Distance & Num. & 7 & No. of other discourse units between unit 1 and unit 2 \\ 
        Lexical Overlap & Num. & 3 & No. of overlapping non-stoplist words in unit 1 and unit 2 \\
    \hline
    \end{tabular}
    \caption{Sample of features used for the relation classification module.
    Asterisked features apply twice for each instance, once for each unit.
    Combination of features varies per corpus---see code for full details.
    }
    \label{tab:relfeats}
\end{table*}

For the plain text segmentation scenario, we generate automatic sentence splits and Universal Dependencies parses using the Transformer-based sentence splitter used in the AMALGUM corpus \cite{GesslerEtAl2020} trained on the treebanked shared task training data, tagged using Stanza \cite{qi-etal-2020-stanza} and parsed using DiaParser\footnote{https://github.com/Unipisa/diaparser} \cite{AttardiEtAl2021}. 
For \model{fas.rst.prstc} and \model{zho.rst.sctb}, we split the text based on punctuation (on ‘.’, ‘!’, ‘?’ and Chinese equivalents) since experiments revealed that this approach yields better sentence boundaries.

\subsection{Relation Classification}

Our relation classification module has a simple architecture: a pretrained BERT model is used (again varying by language---cf.~\Cref{tab:lms}), and a linear projection and softmax layer is used on the output of the pooling layer to predict the label of the relation. 
The two units involved in every relation are prepared just as if they were being prepared for BERT's Next Sentence Prediction (NSP) task: a [CLS] token begins the sequence, a [SEP] token separates the two units in the sequence, and another [SEP] token appears at the end of the sequence. 
As an example, consider this instance from \model{eng.sdrt.stac}:
\[ \texttt{\small[CLS] do we start ? [SEP] no [SEP]} \]
Though this model was originally intended as a baseline, further experiments with e.g.~a separate encoder proved to be much less competitive.

Our exact choice of pretrained model differs in most cases from the one used in the segmentation and connective detection task, primarily due to superior performance by models that were pretrained using the NSP task and had a pretrained pooler layer. 
This restricts the LM choice: for example, most models that are styled after RoBERTa \citep{liu2019roberta} are not pretrained using an NSP task. 
We select models using the same process as before, based on optimal performance on the validation (dev) sets of the corpora.


The system is further enhanced with features. First, the direction feature on each relation is encoded using pseudo-tokens: if the direction of the relation is left to right (1>2), we insert the tokens {\tt \}} and {\tt >} around the first unit. 
In the example above, the direction of the relation is left to right (1>2), and the resulting sequence with pseudo-tokens is:

\[ \texttt{\small[CLS] \} do we start ? > [SEP] no [SEP]} \]
The same is done for right-to-left units, where the characters {\tt \{} and {\tt <} are used instead, but surrounding the second unit:
\[ \texttt{\small[CLS] thanks [SEP] < im ok \{ [SEP]} \]
Our motivation in doing this is to represent directionality for the BERT encoder in its native feature space, and experimental data show that it is helpful.

Second, we introduce hand-crafted features in a step between the BERT model's embedding and encoder layers. Recall that BERT has a static embedding layer which projects each word-piece into its initial vector representation.
Just before this input is sent to the Transformer encoder blocks, we expand the sequence by inserting a new vector in between the [CLS] token and the first token of unit 1.
This feature vector bears sequence-level information, where categorical and numerical features have been encoded into a vector just as for the segmentation and connective detection module: numerical features are optionally log scaled or binned and embedded, and categorical features are embedded. 
The remaining dimensions after all features have been added to the vector are padded with 0. 

Unlike our approach for segmentation and connective detection, we change which features we use on a per-corpus basis, as preliminary experiments showed that using all features for all corpora can produce significant degradations, which we hypothesize are caused by feature sparseness in the training split leading to overfitting. 
A sample of the features we used is in \Cref{tab:relfeats}, and the list of which features were used for which corpus is available in our code. 

Specifically, for the \textsc{Lexical Overlap} feature in the table, we used the freely available stoplists used by the Python library spaCy \citep{spacy}. The \textsc{Same Speaker} feature has proven very useful in the STAC dataset, which is a corpus of strategic chat conversations \cite{asher-etal-2016-discourse}. The \textsc{Distance} feature is used in half of the datasets and has shown effectiveness regardless of annotation framework. Similarly, the \textsc{Position} feature has been shown to be beneficial for half of the corpora. The \textsc{Length Ratio} feature proved to be effective for the three PDTB-style datasets. For RST-style corpora, the number of \textsc{Children} of a nucleus or satellite unit is more effective. Moreover, the \textsc{Discontinuous} feature has also contributed to performance gain in several RST-style corpora such as \model{eng.rst.gum}, \model{eng.rst.rstdt}, \model{por.rst.cstn}, \model{spa.rst.rststb}, \model{zho.rst.sctb}, and \model{fas.rst.prstc}. The \textsc{Genre} feature is beneficial in corpora that have a wide range of text types such as \model{eng.rst.gum}. The direction feature was also included in the feature vector, as experiments showed that including it was helpful, despite the fact that the pseudo-tokens were already expressing it to the BERT encoder.



\section{Results}

\def\P{\phantom{$-$}}
\begin{table*}[ht]%
\centering
\begin{subtable}{0.3985\textwidth}
\resizebox{\columnwidth}{!}{%
\begin{tabular}{l|c|c|c|l|c}
\hline
\textbf{Corpus} & \textbf{P} & \textbf{R} & \textbf{F$_1$} & \textbf{2019 Best F$_1$} & \textbf{vs. 2019} \\
\hline
deu.rst.pcc      & 97.07 & 94.15 & \textbf{95.58} & 94.99  (ToNy) & \P0.59\P \\
eng.rst.gum      & 93.90 & 94.43 & \textbf{94.15} & --  & -- \\
eng.rst.rstdt    & 96.39 & 96.89 & \textbf{96.64} & 96.04 (ToNy) & \P0.60\P \\
eng.sdrt.stac    & 96.25 & 93.63 & 94.91 & \textbf{95.32}  (GumDrop) & $-$0.41\P \\
eus.rst.ert      & 93.42 & 87.73 & \textbf{90.46} & --  & -- \\
fas.rst.prstc    & 92.79 & 93.10 & \textbf{92.94} & --  & -- \\
fra.sdrt.annodis & 89.43 & 90.65 & \textbf{90.02} & -- & -- \\
nld.rst.nldt     & 97.50 & 94.50 & \textbf{95.97} & 95.45 (GumDrop) & \P0.52\P \\
por.rst.cstn     & 93.18 & 95.56 & \textbf{94.35} & 92.92 (ToNy) & \P1.43\P \\
rus.rst.rrt      & 85.57 & 86.89 & \textbf{86.21} & -- & -- \\
spa.rst.rststb   & 92.53 & 91.96 & \textbf{92.22} & 90.74 (ToNy) & \P1.48\P \\
spa.rst.sctb     & 83.44 & 81.55 & 82.48 & \textbf{83.12} (ToNy) & $-$0.64\P \\
zho.rst.sctb     & 90.30 & 77.38 & \textbf{83.34} & 81.67 (DFKI) & \P1.67\P \\
\hline
eng.pdtb.pdtb & 92.93 & 91.15 & \textbf{92.02} & -- & \P--\P \\
tur.pdtb.tdb  & 93.71 & 94.53 & \textbf{94.11} & -- & \P--\P \\
zho.pdtb.cdtb & 89.19 & 85.95 & \textbf{87.52} & -- & \P--\P \\
\hline
\textbf{mean} & 92.35 & 90.63 & 91.43 & -- & -- \\
\hline
\end{tabular}
}    
\caption{Results for Gold Treebanked Data.}
\label{tab:seggold}
\end{subtable}%
\hspace{0.04\textwidth}%
\begin{subtable}{0.4615\textwidth}%
\resizebox{\columnwidth}{!}{%
\begin{tabular}{l|c|c|c|l|c|c}
\hline
\textbf{Corpus} & \textbf{P} & \textbf{R} & \textbf{F$_1$} & \textbf{2019 Best F$_1$} & \textbf{vs. 2019} &  \textbf{vs. Gold} \\
\hline
deu.rst.pcc      & 95.15 & 92.86 & 93.94 & \textbf{94.68} (ToNy)    & $-$0.74\P & $-$1.64 \\
eng.rst.gum      & 92.65 & 92.59 & \textbf{92.61} & --              &  --     & $-$1.54\\
eng.rst.rstdt    & 96.80 & 95.92 & \textbf{96.35} & 93.43 (ToNy)    &  \P2.92\P & $-$0.28 \\
eng.sdrt.stac    & 91.77 & 92.06 & \textbf{91.91} & 83.99 (ToNy)    &  \P7.92\P & $-$3.00 \\
eus.rst.ert      & 92.70 & 88.38 & \textbf{90.47} & --              &  --     & \phantom{$-$}0.01\\
fas.rst.prstc    & 92.95 & 92.78 & \textbf{92.86} & --              &  --     & $-$0.08\\
fra.sdrt.annodis & 87.95 & 83.79 & \textbf{85.78} & --              &  --     & $-$4.24\\
nld.rst.nldt     & 96.97 & 92.54 & \textbf{94.69} & 92.32 (ToNy)    &  \P2.37\P & $-$1.29\\
por.rst.cstn     & 93.21 & 95.03 & \textbf{94.11} & 91.86 (ToNy)    &  \P2.25\P & $-$0.25\\
rus.rst.rrt      & 87.31 & 84.24 & \textbf{85.74} & --              &  --     & $-$0.47\\
spa.rst.rststb   & 93.30 & 90.30 & \textbf{91.76} & 89.60 (ToNy)    &  \P2.16\P & $-$0.46\\
spa.rst.sctb     & 83.97 & 77.98 & 80.86 & \textbf{81.65} (ToNy)    & $-$0.79\P & $-$1.62\\
zho.rst.sctb     & 84.04 & 70.00 & \textbf{76.21} & 73.10 (GumDrop) &  \P3.08\P & $-$7.13\\
\hline
eng.pdtb.pdtb    & 94.29 & 90.92 & \textbf{92.56} & -- &  \P--\P & \phantom{$-$}0.54 \\
tur.pdtb.tdb     & 91.98 & 95.22 & \textbf{93.56} & -- &  \P--\P & $-$0.55 \\
zho.pdtb.cdtb    & 90.27 & 86.54 & \textbf{88.35} & -- &  \P--\P & \phantom{$-$}0.83 \\
\hline
\textbf{mean}    & 91.58 & 88.82 & 90.11 & -- & -- & $-$1.32 \\
\hline
\end{tabular}
}
\caption{Results for Plain Tokenized Data.}
\end{subtable}
\caption{Segmentation and Connective Detection Results.
All numbers are averaged over five runs in order to accommodate instability in the training process which leads to varying performance.
If a corpus was included in the 2019 shared task and has not been significantly modified since then, we also include the best result on the corpus in 2019 for comparison. }%
\label{tab:seg}%
\end{table*}

\subsection{Segmentation and Connectives}
Table \ref{tab:seg} gives scores on EDU segmentation and connective detection in the two shared task scenarios: treebanked and plain text, as well as the best previously reported score and system for datasets which are unchanged from 2019 (see \citealt{zeldes-etal-2019-disrpt} for details). 
We find strong performance in both the treebanked and plain tokenized data scenarios: our system nearly always outperforms the best score from 2019, and we observe especially large gains for connective detection.

On treebanked data, the results show that performance has improved since 2019 on nearly all unchanged datasets, with degradations of only around 0.5\% for \model{eng.sdrt.stac} and \model{spa.rst.sctb} compared to the previous best systems, GumDrop and ToNy respectively. 
For some datasets, gains are dramatic, most notably for Turkish (14.5\% gain) and Chinese connective detection (8.4\%), which is perhaps due to the availability of better language models and our use of conditional random fields. On average the improvement on treebanked data is close to 3\% for datasets represented in 2019.

On plain tokenized data, the improvement from 2019 is even more pronounced, with an average gain of 3.7\% compared to 2.8\% for treebanked data.
While performance on some corpora was roughly constant regardless of whether data was treebanked or plain tokenized (e.g. \model{eng.rst.rstdt}, \model{por.rst.cstn}), it dropped considerably for some corpora on plain tokenized data.
This effect is most dramatic for \model{zho.rst.sctb}, where we see a degradation of 7.1\%. 
This effect cannot be explained just by the amount of training data available: correlation between training token count and degradation is low (Pearson's $r$ = 0.092, $p$ = 0.74).

We speculate that these causes of the degradations are primarily due to idiosyncrasies of the corpora.
\model{eng.sdrt.stac}, for instance, has a mild degradation ($-$3\%), which we believe to be primarily due to the lack of punctuation and capitalization compared to e.g.~a newswire corpus like \model{eng.rst.rstdt}, which exhibited very little degradation. 
From this we might infer that degradation in the absence of treebanked data will be correlated with the degree to which predicting sentence splits from plain text is difficult.
We additionally hypothesize that a lack of gold sentence breaks affects RST datasets more than PDTB datasets, since the beginning of a sentence is almost always the beginning of a new elementary discourse unit, while connectives are mainly identified lexically, and need to be identified regardless of the relative position of sentence splits.

\paragraph{Ablation Study}

\begin{table}
\resizebox{\columnwidth}{!}{%
\begin{tabular}{lccccc}
\hline
\textbf{Corpus} & \textbf{F$_1$} (all) & \multicolumn{2}{c}{\textbf{F$_1$} (no feats.)} & \multicolumn{2}{c}{\textbf{F$_1$} (CWE only)} \\
& & {\it abs.} & {\it gain} & {\it abs.} & {\it gain} \\
\hline
deu.rst.pcc      & 95.58 & 96.28 &  -0.70            & 95.81 & -0.23            \\
eng.rst.gum      & 94.15 & 89.74 &  \hphantom{0}4.41 & 92.23 & \hphantom{0}1.92 \\
eng.rst.rstdt    & 96.64 & 91.41 &  \hphantom{0}5.23 & 94.59 & \hphantom{0}2.05 \\
eng.sdrt.stac    & 94.91 & 94.54 &  \hphantom{0}0.37 & 94.54 & \hphantom{0}0.37 \\
eus.rst.ert      & 90.46 & 91.01 &  -0.54            & 90.76 & -0.30            \\
fas.rst.prstc    & 92.94 & 92.99 &  -0.05            & 93.24 & -0.30            \\
fra.sdrt.annodis & 90.02 & 88.80 &  \hphantom{0}1.22 & 88.48 & \hphantom{0}1.54 \\
nld.rst.nldt     & 95.97 & 95.48 &  \hphantom{0}0.50 & 95.11 & \hphantom{0}0.86 \\
por.rst.cstn     & 94.35 & 92.65 &  \hphantom{0}1.71 & 93.93 & \hphantom{0}0.42 \\
rus.rst.rrt      & 86.21 & 86.18 &  \hphantom{0}0.03 & 86.01 & \hphantom{0}0.20 \\
spa.rst.rststb   & 92.22 & 92.04 &  \hphantom{0}0.18 & 92.39 & -0.17            \\
spa.rst.sctb     & 82.48 & 84.21 &  -1.73            & 83.32 & -0.84            \\
zho.rst.sctb     & 83.34 & 84.87 &  -1.54            & 82.60 & \hphantom{0}0.74 \\
\hline                    
eng.pdtb.pdtb    & 92.02 & 87.72 &  \hphantom{0}4.30 & 82.42 & \hphantom{0}9.60 \\
tur.pdtb.tdb     & 94.11 & 94.02 &  \hphantom{0}0.10 & 93.54 & \hphantom{0}0.57 \\
zho.pdtb.cdtb    & 87.52 & 88.53 &  -1.01            & 88.14 & -0.62            \\
\hline                        
\textbf{mean}    & 91.43 & 90.65 &  \hphantom{0}0.78 & 90.45 & \hphantom{0}0.99 \\
\hline
\end{tabular}
}
\caption{F$_1$ scores for ablations on gold treebanked data: next to normal scores from \Cref{tab:seggold}, we report scores without handcrafted features, and without character embeddings and fastText static word embeddings, as well as the ``gain'' for each (non-ablated score -- ablated score). Due to time constraints, ablations are based on {\it three} runs instead of the standard five.}
\label{tab:seg_abl}%
\end{table}

In order to assess the importance of the various modules of our segmentation and connective detection system, we conduct an ablation study.
Due to the large computational expense of conducting full runs over all datasets, we choose only two ablation conditions.
In the first, we remove all handcrafted features described in \Cref{tab:segfeats}.
In the second, we remove character embeddings and fastText static word embeddings, leaving only contextualized word embeddings. 
The results of this study is given in \Cref{tab:seg_abl}.

The general trend in the results of the ablation study seems to be that both handcrafted features and supplementary word embeddings are helpful on average, though they may sometimes lead to minor degradations, and have a dramatically pronounced effect on a few corpora in particular.
Handcrafted features have a mild effect on most corpora but lead to large gains for GUM, RST-DT, and PDTB. 
It is not immediately clear why this might be: performance on GUM, which is diverse with respect to genre, probably benefits from having a genre feature, but RST-DT and PDTB are homogeneous with respect to genre. We also note that GUM, RST-DT, and especially PDTB are large corpora, so perhaps the explanation lies in their size, but RRT is also very large and has multiple genres, yet handcrafted features led to nearly no gain on this dataset. 

Turning now to the CWE-only ablation, we see a similar pattern: most corpora are only minorly affected by the inclusion of non-CWE embeddings, with a couple (GUM, RST-DT) showing a moderate gain of 2\%, and one corpus (PDTB) showing an anomalous gain of 10\%. 
Just as with the handcrafted feature ablation, it is difficult to know what could explain these corpora's divergent behavior. Ordinarily, static word embeddings might benefit small corpora with OOV items in the test set, since the embedding space will be stable in the unseen data -- however PDTB is very large and homogeneous (newswire), making this explanation unlikely.
Since no other corpus showed such a dramatic drop with non-CWE embeddings ablated, and since other CWE-based systems at DISRPT 2021 score around what our system would have scored if the drop had been more in line with what was observed for other corpora (2\% drop, for a score in the low 90s, as was achieved by disCut and SegFormers), we speculate that the 10\% drop observed here is due to some kind of implementation error or statistical fluke due to the nondeterminism of training models on GPUs, though the effect survives in the system reproduction on the Shared Task evaluators' machine.

In sum, our ablation study for segmentation and connective detection suggests that both handcrafted features and non-CWE embeddings are not silver bullets, though they are often helpful.
Degradations were seen more often on smaller datasets, which perhaps indicates that in low-data situations these additional resources can serve more than anything as a source of overfitting. 
But both were on average responsible for a 1\% gain,\footnote{Assuming that they can be treated independently, which is an idealization.} which shows they are both useful, and invites the question of whether there might be even better handcrafted features, which could be tailored more accurately to properties of specific target languages and genres.

\subsection{Relation Classification}

\begin{table}[t]
\resizebox{\columnwidth}{!}{%
\begin{tabular}{lcccc}
\hline
\textbf{Corpus}  & \textbf{\# Relations} & \twoline{\textbf{Accuracy}}{\textbf{(w/ feats.)}} & \twoline{\textbf{Accuracy}}{\textbf{(w/o feats.)}} & \twoline{\textbf{Feature}}{\textbf{Gain}} \\\hline
deu.rst.pcc      & 26 & 39.23 & 33.85 & \phantom{0}5.38 \\
eng.pdtb.pdtb    & 23 & 74.44 & 75.63 & -1.19 \\
eng.rst.gum      & 23 & 66.76 & 62.65 & \phantom{0}5.55 \\
eng.rst.rstdt    & 17 & 67.10 & 66.45 & \phantom{0}0.65 \\
eng.sdrt.stac    & 16 & 65.03 & 59.67 & \phantom{0}5.36 \\
eus.rst.ert      & 29 & 60.62 & 59.59 & \phantom{0}1.03 \\
fra.sdrt.annodis & 18 & 46.40 & 48.32 & -1.92 \\
nld.rst.nldt     & 32 & 55.21 & 52.15 & \phantom{0}3.06 \\
por.rst.cstn     & 32 & 64.34 & 67.28 & -2.94 \\
rus.rst.rrt      & 22 & 66.44 & 65.46 & \phantom{0}0.98 \\
spa.rst.rststb   & 29 & 54.23 & 54.23 & \phantom{0}0.00 \\
spa.rst.sctb     & 25 & 66.04 & 61.01 & \phantom{0}5.03 \\
tur.pdtb.tdb     & 23 & 60.09 & 57.58 & \phantom{0}2.51 \\
zho.pdtb.cdtb    & 9  & 86.49 & 87.34 & -0.85 \\
zho.rst.sctb     & 26 & 64.15 & 64.15 & \phantom{0}0.00 \\
fas.rst.prstc    & 17 & 52.53 & 51.18 & \phantom{0}1.35 \\\hline
\textbf{mean}    &    & 61.82 & 60.41 & \phantom{0}1.41 \\
\hline
\end{tabular}
}
\caption{Relation Classification Results. The score for each corpus is averaged over 5 runs. Also included is score {\it without} any hand-crafted features.}
\label{tab:rel-clf-results}
\end{table}

Table \ref{tab:rel-clf-results} gives scores (averaged over 5 runs for each corpus) on relation classification on all 16 corpora. 
We include performance on all corpora without any hand-crafted features added in order to assess their utility, and we find that they appreciably boost performance, granting on average a 1.5\% accuracy gain, with some of the biggest gains on small corpora with many labels like \model{deu.rst.pcc} ($+$5.38\%) and \model{spa.rst.sctb} ($+$5.03\%).
Since difficulty of classification increases with number of labels, we also include the number of relation types for each corpus in order to contextualize the scores.
The \model{zho.pdtb.cdtb} corpus achieved the highest accuracy score, as there are only 9 relation types to classify,%
\footnote{Unlike the other two PDTB-style corpora (i.e.~\model{eng.pdtb.pdtb} and \model{tur.pdtb.tdb}), where the predicted labels are truncated at Level-2 (e.g.~\textsc{Temporal.Asynchronous}), the relation labels in \model{zho.pdtb.cdtb} only contain one level (e.g.~\textsc{Temporal}).} %
and scores tend to be lower for corpora with many relations like \model{nld.rst.nldt}.

There is much variance in how much performance on each corpus was able to benefit from having additional features.
Many of the corpora that had the largest gains are small, but this is not always the case: \model{tur.pdtb.tdb}, one of the larger corpora, has its score improved by 2.5\%. 
On the other hand, while small corpora generally seem to benefit more from features, not all are able to: \model{fra.sdrt.annodis}, a small corpus, sees a degradation of 2\% with features.
We expect that much of the differential benefit of features is to be explained by the nature of the label-sets used in different corpora, and the available features. 
No two of these corpora use the exactly same label set, and label sets vary quite a bit in the linguistic phenomena that they encode and are sensitive to. 

Additionally, different corpora have different features available, such as genre (GUM, RRT, RSTSTB, PRSTC, SCTB -- yes; PCC, PDTB, RST-DT, TDB -- no), gold speaker information (STAC), discontinuity (Annodis, ERT, GUM, PDTB, RST-DT -- yes; PCC, NLDT, STAC -- no) etc., meaning that looking at gain across datasets is not comparing like with like. While some features are available for all corpora, such as distance, unit lengths, or position in document, others are restricted by framework, such as number of children, which is not relevant for PDTB-style data. A set of formalism-agnostic features (e.g.~length\_ratio, is\_sentence, and the direction of the dependency head of the unit) were used for PDTB-style data across the board and were only effective for TDB: we hypothesized that the English PDTB dataset is so big that generic features do not add much value; for CDTB, as we found in our error analysis, the 9 relations sometimes are not very distinct from each other, and these generic features do not help with disambiguation in those cases. 

Overall, a picture emerges with relations that is similar to the one that arose with segmentation and connective detection: features are helpful on the whole, slightly harmful in some cases, and especially helpful on some corpora. More work remains to be done in understanding the contribution of individual features and how these relate to the frameworks and data types available in each language.

\section{Discussion}


\begin{figure*}%
    \centering
    \begin{subfigure}{0.48\textwidth}
    \includegraphics[width=9cm, trim={0cm 0.5cm 1cm 0cm}, clip]{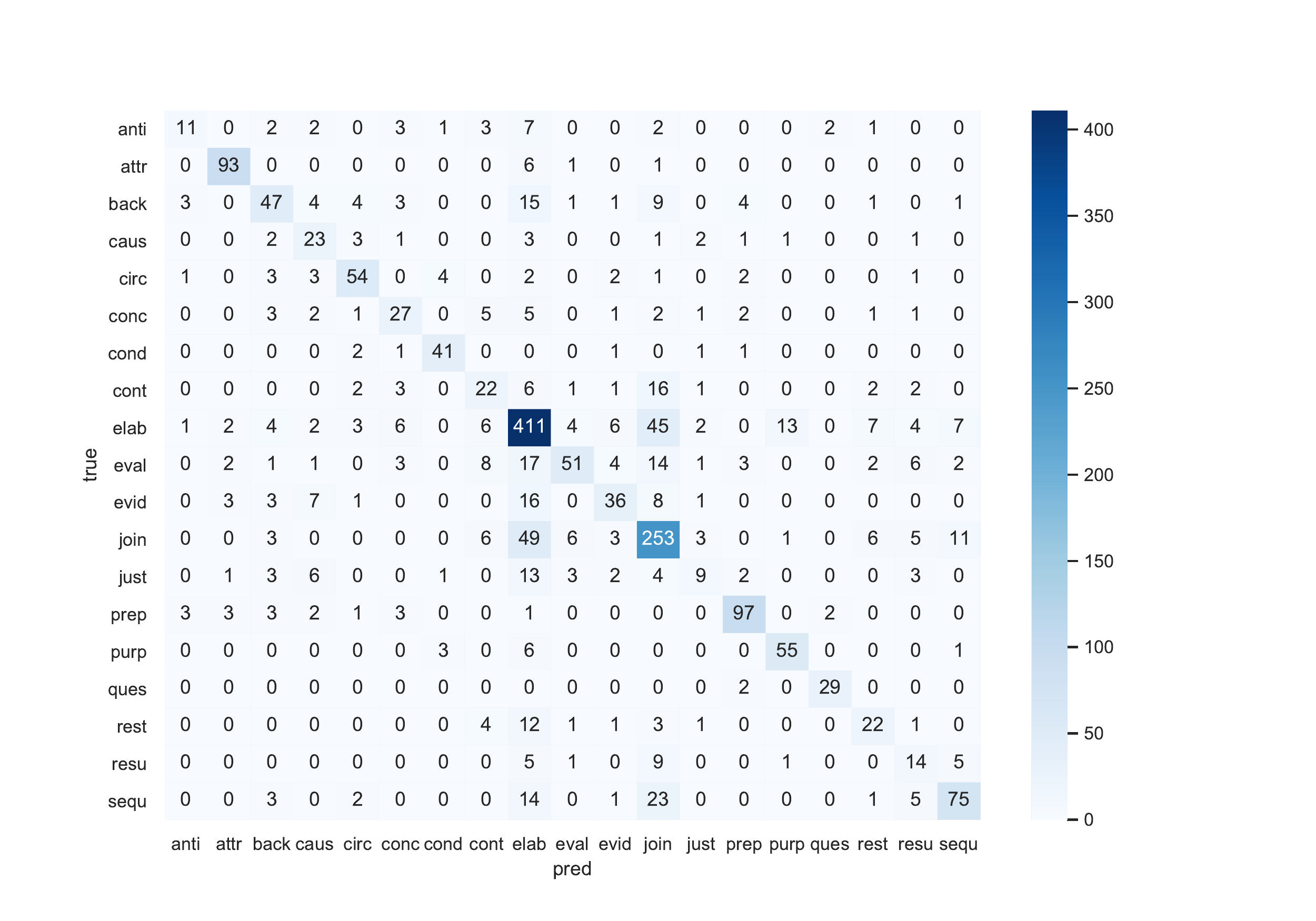}
    \caption{eng.rst.gum (GUM)}
   \end{subfigure}
    \begin{subfigure}{0.48\textwidth}
    \includegraphics[width=9cm, trim={0cm 0.5cm 1cm 0cm}, clip]{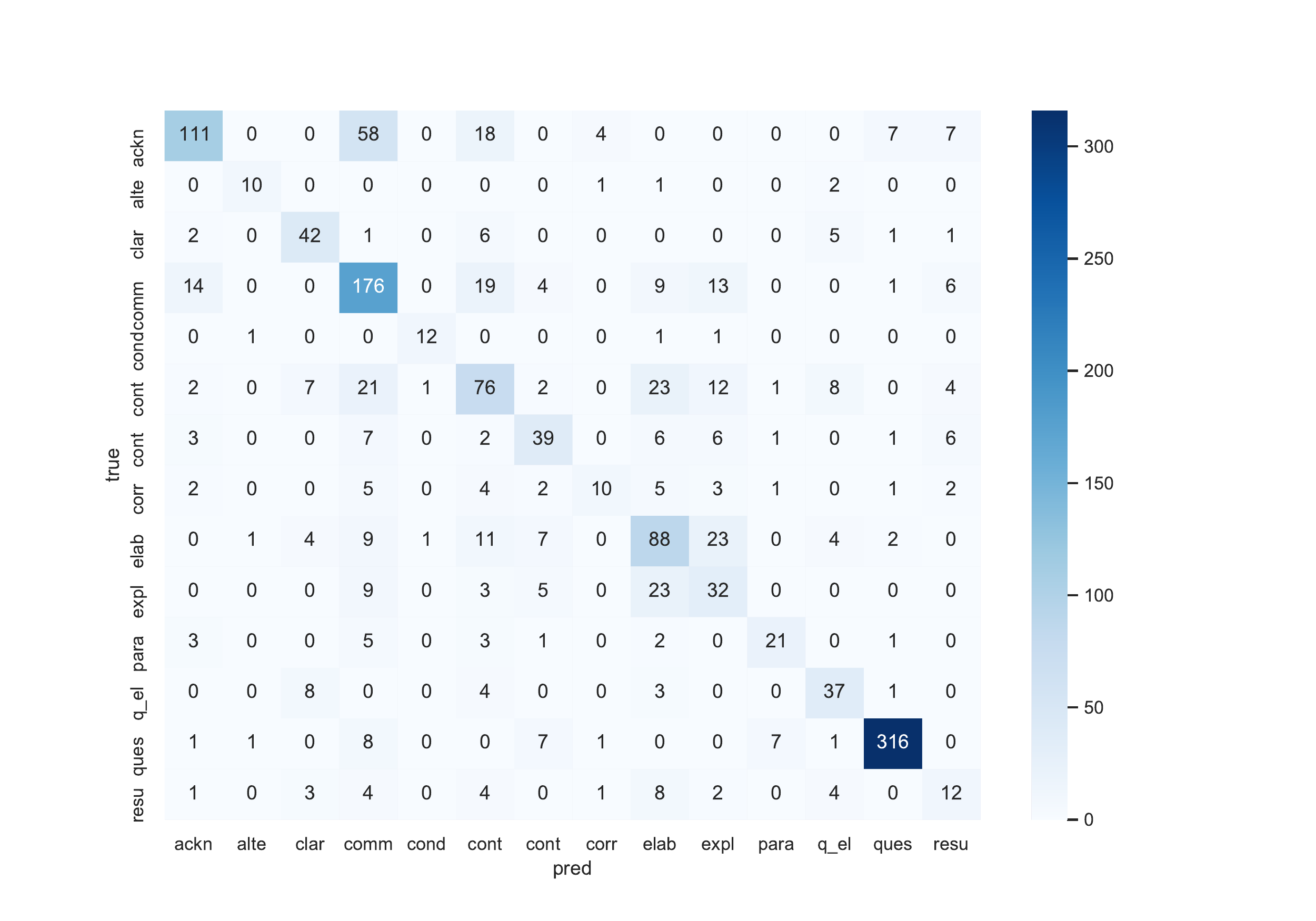}
    \caption{eng.sdrt.stac (STAC)}
\end{subfigure}
    \caption{Confusion Matrices for Common Relations in the Highest and Lowest Scoring EDU Datasets. }%
    \label{fig:confmat}%
\end{figure*}

Figure \ref{fig:confmat} shows confusion matrices for common relations in the highest and lowest scoring EDU datasets, \model{eng.rst.gum} (GUM) and \model{eng.sdrt.stac} (STAC). 
Both panels reveal issues with over-prediction of the most common labels, which can be thought of as `defaults': the most common \textsc{elaboration} in GUM and the second most common, \textsc{Comment} in STAC. 
The actual most common relation in STAC, \textsc{Question}, does not suffer from false positives, likely due to a combination of the frequent and reliable question mark cue, \textit{wh}-words or subject-verb inversion, combined with the availability of gold speaker and direction information (\textsc{Question} only links units from different speakers, left to right). 
The same is true for \textsc{question} in GUM, which also obtained a comparatively high score. 
Conversely, rarer relations are hardly predicted, with \textsc{antithesis} in GUM being predicted for less than half of its true instances, and similarly for \textsc{Result} in STAC, suggesting a class imbalance problem, in particular given that both these relations are sometimes marked by overt discourse markers.

Although EDU datasets (RST/SDRT) do not distinguish explicit and implicit relations, analysis suggests that explicit signals are important. For GUM, the model scored high on medium-frequency relations with clear cues such as \textsc{attribution}, which are always signalled by attribution verbs such as \textit{believe} and \textit{say}. This also holds true for the \model{eng.rst.rstdt} corpus where \textsc{attribution} is the highest-scoring relation. GUM's \textsc{elaboration} and \textsc{joint} as well as \textsc{sequence} and \textsc{joint} are two pairs of relation labels that are most frequently confused with each other, as the former pair contains the two labels that are both overgeneralized and the latter contains the two labels that are both \textsc{multinuclear} relation types that could be confusing when no explicit connective or lexical item indicating a sequential order of actions or events is present \cite{liu-2019-beyond}. Relations with relatively unambiguous markers, such as \textsc{condition}, show good results in both GUM and STAC, indicating that even relatively rare relations can be identified if they are usually explicitly signalled.

Relations such as \textsc{justify}, \textsc{result}, and \textsc{cause} scored low in both matrices as such instances in the test data often do not have explicit discourse markers to help understand the rhetorical relation between the units in context. In the presence of an ambiguous discourse marker, predictions prefer the relation that is more prototypically associated with that marker: for instance, the gold relation label for \ref{gum_error_1} is \textsc{result} whereas the model classified it as \textsc{sequence}, likely due to the fact that the discourse marker \textit{then} tends to be a strong signal for \textsc{sequence} indicating something sequential is involved. 

\ex. \textbf{Then} she lets go and falls . You scream . [GUM\_fiction\_falling] \label{gum_error_1}

In fact, if it were not the case that RST mandates a single outgoing relation per discourse unit, it would be possible to claim a concurrent sequential relation (likely to a previous unit) next to the annotated \textsc{result} relation for the pair in the example (see \citealt{Stede2008} on concurrent relations in RST).

\section{Conclusion}

We have presented \sysname, a system for all tasks in the DISPRT 2021 Shared Task: EDU segmentation, connective detection, and relation classification. 
Our system relies on sequence tagging and sentence pair classification architectures powered by CWEs and supported by rich, handcrafted, instance-level features, such as position in the document, distance between units, gold speaker information, document metadata, and more. 

Our results suggest that powerful pretrained language models are the main drivers of performance, with additional features providing small to medium improvements (with some exceptions, such as the high importance of speaker information for chat data as in STAC). 
For relation classification, CWEs pretrained using an NSP task proved to be superior.

Our error analysis suggests unsurprisingly that class imbalances, especially in the case of relations that tend to be implicit (i.e.~lack overt lexical signals), lead to over-prediction of majority classes, suggesting a need for more training data for the minority ones. 
However, we are encouraged by improvements on datasets that were featured in the 2019 Shared Task \cite{zeldes-etal-2019-disrpt}, and the overall high scores obtained by the system across a range of datasets, all while including some correct predictions for relatively rare relations. 
We hope that the growing availability of annotated data, coupled with architectures that can harness pre-trained models, will lead to further improvements in the near future.


\bibliography{custom}
\bibliographystyle{acl_natbib}

%
%

\end{document}